%% file: main_wo_comment.tex
\documentclass{article}
\usepackage[utf8]{inputenc}
\usepackage[square,numbers]{natbib}
\usepackage{pifont}
\usepackage{adjustbox}

\usepackage{verbatim}
\usepackage{todonotes}
\usepackage{amsfonts}
\usepackage{amsmath}
\usepackage{amsthm}
\usepackage{algorithm}  
\usepackage{algorithmic}
\usepackage{mathtools}
\usepackage{thmtools} 
\usepackage{thm-restate}
\usepackage{enumitem}
\usepackage{graphicx}
\usepackage{float}
\usepackage{subcaption}

\usepackage[colorlinks=false]{hyperref}
\usepackage{amsmath,amssymb, amsfonts}
\usepackage{bm,bbm}

\theoremstyle{definition}

\usepackage[a4paper, total={6in, 8in}]{geometry}
\newtheorem{definition}{Definition}
\newtheorem{theorem}{Theorem}
\newtheorem{lemma}{Lemma}

\newtheorem{assumption}{Assumption}
\newtheorem{proposition}{Proposition}

\input{paper/definitions}

\author{Apurv Shukla \footnote{apurv.shukla@tamu.edu. This work was primarily done when the author was a graduate student at Columbia University. The author would like to thank Rachel Cummings for providing extensive feedback on earlier versions of this manuscript and advising this project.}}
\title{Differentially Private High-dimensional Bandits}
\date{\today}
\begin{document}
\maketitle

\begin{abstract}
\input{paper/abstract}
\end{abstract}

\section{Introduction}
\input{paper/introduction}

\section{Model and Preliminaries}
\input{paper/framework}

\section{PrivateThresholdLASSO}
\label{sec:algorithm}
\input{paper/threshold-lasso}

\section{Privacy Guarantees}
\label{sec:privacy}
\input{paper/privacy}

\section{Regret Guarantee}
\label{sec:regret}
\input{paper/regret}

\section{Lower Bounds}
\label{sec:lower-bound}
\input{paper/lower-bound}

\bibliographystyle{apalike}
\bibliography{paper/ref}
\appendix
\end{document}

%% file: paper/definitions.tex
\def\cA{\mathcal{A}}
\def\cB{\mathcal{B}}

\def\cD{\mathcal{D}}
\def\cE{\mathcal{E}}
\def\cF{\mathcal{F}}
\def\cG{\mathcal{G}}
\def\cH{\mathcal{H}}

\def\cM{\mathcal{M}}
\def\cN{\mathcal{N}}
\def\cO{\mathcal{O}}
\def\cP{\mathcal{P}}

\def\cR{\mathcal{R}}
\def\cS{\mathcal{S}}
\def\cT{\mathcal{T}}
\def\cU{\mathcal{U}}
\def\cV{\mathcal{V}}
\def\cW{\mathcal{W}}
\def\cX{\mathcal{X}}

\def\mE{\mathbb{E}}

\def\mP{\mathbb{P}}

\def\mR{\mathbb{R}}


%% file: paper/abstract.tex
We consider a high-dimensional stochastic contextual linear bandit problem when the parameter vector is $s_{0}$-sparse and the decision maker is subject to privacy constraints under both central and local models of differential privacy. We present PrivateLASSO, a differentially private LASSO bandit algorithm. PrivateLASSO is based on two sub-routines: (i) a sparse hard-thresholding-based privacy mechanism and (ii) an episodic thresholding rule for identifying the support of the parameter $\theta$. We prove \textit{minimax private} lower bounds and establish privacy and utility guarantees for PrivateLASSO for a central model under standard assumptions. 

%% file: paper/introduction.tex
The stochastic contextual linear bandit problem considers a decision maker who is given an exogenous set of contexts and then chooses an action from a finite set and receives a reward. The decision maker's objective is to minimize its regret (equivalently, maximize reward). To do this, the decision maker faces an exploration-exploitation trade-off, wherein it explores the action space to find the optimal action while exploiting the information accumulated to minimize regret. The observed rewards depend linearly on the observed contexts and only on a small subset of the coordinates, resulting in a \emph{sparse} estimation problem. 

To motivate the need for privacy in this setting, consider the clinical trial setting where the decision maker's goal is to find the safety-efficacy level of a given pharmaceutical drug across different patient populations. A patient arrives for a clinical trial with covariates describing their demographics, and the clinic responds with drug dosage levels specific to the patient's covariates. The patient covariates thus map the drug-dosage levels to a feature vector
and the expected reward depends on
the same unknown linear function of the covariates. Not all covariates are necessary for learning the drug safety-efficacy curve. In fact, often a small subset of covariates is sufficient for learning, suggesting that the parameter vector is sparse. Further, the patient's covariates are sensitive personal information but are still necessary for learning the appropriate drug dosage across multiple populations. 

Differential privacy~\cite{DMNS06} has become the gold standard for privacy-preserving data analysis. It requires that a single data record has only a bounded effect on the distribution of an algorithm's output. In our bandit setting, we consider a relaxed variant known as \emph{joint differential privacy}~\cite{KPRU14} which ensures that the $i$th data record has a bounded effect on the joint distribution of all outputs besides the $i$th one. This allows, for example, the treatment of patient $i$ to depend arbitrarily on her own covariates, while still maintaining differential privacy concerning the data of other patients.

In this paper, we consider the stochastic linear contextual bandit problem when the underlying parameter vector is sparse, i.e., the ambient parameter dimension $d \gg T$ and the parameter vector is  $s_0$-sparse. Our goal is to provide algorithms that preserve the privacy of the patient covariates $\cX_t$ while achieving regret that is sub-linear in $T$, sub-polynomial in $d$ and polynomial in $s_0$.

\subsection{Our Contributions} 
In this paper, we provide a jointly differentially private algorithm, PrivateLASSO (Algorithm~\ref{algo:private-lasso}), for the LASSO bandit problem. This algorithm builds upon a non-private thresholding algorithm for the LASSO bandit problem~\cite{thlasso}, and privatizes the approach using a variant of the SparseVector Technique~\cite{DNRRV09,dwork2010differential} and tree-based aggregation~\cite{chan2011private,dwork2010differential}.  The algorithm first estimates the $s_0$-sparse support of $\theta$ using the SparseVector variant to privately identify which coordinates of a non-private $\ell_1$-regression estimate $\hat{\theta}$ were above a threshold, and thus likely to be non-zero in the true vector $\theta$. It then performs an $\ell_2$ regression on only the identified support, using tree-based aggregation to avoid redoing the regression at every time, and thus avoiding large privacy composition. We then perform privacy and regret analysis of the PrivateLASSO algorithm to show that it is differentially private (Theorem \ref{thm:private-lasso}) and it achieves regret $\cO( \frac{s^{3/2}_0 \log^3 T}{\varepsilon})$ under the margin condition (Theorem \ref{thm:regret-margin}) or $\cO( \frac{s^{3/2}_{0}\sqrt{T}\log^2 T}{\varepsilon})$ without the margin condition (Theorem \ref{thm:regret-without-margin}). 

\subsection{Related Work}
Multi-armed bandits have been studied since the foundational work of ~\cite{lai1985asymptotically}.  Low-dimensional bandits where the (unknown) underlying parameter vector $\theta$ has dimension $d \ll T$ have been extensively studied; commonly used approaches include upper confidence bounds (UCB)~\cite{agrawal1995sample,auer2002finite} and Thompson Sampling~\cite{agrawal2013thompson,hamidi2020general} (see~\cite{lattimore2020bandit} for a comprehensive overview). In the low-dimensional linear bandit setting, state-of-the-art non-private algorithms attain regret guarantees that scale as $\cO(dK\sqrt{T} )$ for UCB-based policies and $\cO(
d^{3/2}K\sqrt{T})$ for Thompson Sampling based algorithms. 

With the recent rise of differential privacy, privacy-preserving low-dimensional bandit algorithms have been developed recently~\cite{ren2020multi,zheng2020locally,chen2021privacy,han2021generalized}. \cite{mishra2015nearly} propose differentially private UCB and Thompson Sampling-based algorithms for classical $K$-armed bandits with arm-gap $\Delta$ while protecting the privacy of the rewards. The regret scales as $\cO(\frac{K \log^2 T}{\Delta\varepsilon})$ for their UCB-based algorithm and $\cO(\frac{K \log^3 T}{\Delta^2 \varepsilon^2})$ for their Thompson Sampling based algorithm.
\cite{chen2021privacy} propose an algorithm for the linear bandit problem whose regret scales as $\cO(dK\sqrt{T}+\frac{d^2K}{\varepsilon})$. Straightforward extensions of UCB and Thompson Sampling-based algorithms lead to regret guarantees that scale super-linearly with $d$. This motivates the development of algorithms that are based on $\ell_1$-regression and support set estimation. In the non-bandit setting, near-optimal private LASSO algorithms were developed by~\citep{talwar2015nearly}. They propose a Frank-Wolfe-based algorithm with upper and lower-bound guarantees. However, extensions of this algorithm to the bandit setting are non-trivial. The closest to our work is the paper by~\citep{shariff2018differentially}. 
They consider the contextual bandit problem in low dimensions and design an algorithm based on UCB and Tree-based aggregation techniques. The regret of their algorithm scales as $\cO(\frac{d^{3/4}K\sqrt{T}}{\sqrt{\varepsilon}})$. However, since we consider a high-dimensional bandit problem our algorithm design is much more involved and therefore leads to a different regret upper and lower bound analysis. 

\medskip
In this work, we consider the high-dimensional setting ($d \gg T$), so our results are incomparable to the results in the low-dimensional setting. The design of these private algorithms incorporates specialized mechanisms that construct estimators of the mean reward by injecting well-calibrated noise to preserve the privacy of the input data. In almost all cases, the variance of noise -- and therefore the error in the constructed estimate -- scales linearly with the dimension of the parameter. A straightforward adaptation of these tools to the high-dimensional problem setting where $d\gg T$ would lead to a regret bound that scales $\cO(d)$, which would implicitly be super-linear in $T$. Another approach to designing private algorithms is objective perturbation~\cite{chaudhuri2011differentially,kifer2012private}. In the high-dimensional bandit case, a straightforward idea would be to consider objective-perturbation of the $\ell_1$-regularized problem to construct the LASSO estimate in an online fashion, using algorithms developed by, e.g.,~\cite{oh2021sparsity,bastani2020online,thlasso} as off-the-shelf routines. However, algorithms based on such perturbation schemes would again lead to the addition of noise whose variance scales linearly with the dimension $d$, leading to regret super-linear in $T$. 
Our work considers a central model of differential privacy where the decision-maker is responsible for collecting non-private information and publishing private statistics as opposed to a local privacy model wherein the information provided to the decision-maker has been privatized. In the non-bandit setting, the closest work is by~\cite{wang2018high,wang2019sparse} which considers the problem of locally differentially private sparse linear regression. Their algorithm is also based on a hard-thresholding technique in a non-contextual full-information setting.

In the non-private high-dimensional setting, there are several algorithms for the high-dimensional bandit problem~\cite{li2021simple,hao2020high,carpentier2012bandit}, under the necessary assumption that the unknown parameter $\theta$ is $s_0$ sparse. Similar to the arm-separation assumption in the case of classical bandits, this problem has been studied with and without the \emph{margin condition}, that implies that the arm payoffs are sufficiently separated for different context vectors.  \cite{bastani2020online} proposed the LASSO bandit problem and a forced-exploration-based algorithm. Assuming the margin condition, they establish regret guarantees that scale as $\cO(s^2_0(\log d + \log T)^2)$. \cite{oh2021sparsity} propose an algorithm for the LASSO bandit that is agnostic to the sparsity index $s_0$ and whose regret scales as $\cO(s_0\sqrt{T\log dT})$ without the margin condition. Recent work by~\cite{thlasso} proposes a thresholding-based algorithm whose regret is upper bounded by $\cO(s^2_0\log d + s_0\log T)$ assuming the margin condition, and by $\cO(s^2_0\log d + s_0\sqrt{T})$ in the general case.

In this paper, we propose a privacy-preserving algorithm for the high-dimensional $s_0$-sparse LASSO contextual bandit problem, assuming the knowledge of $s_0$. This knowledge of $s_0$ is needed for calibrating noise in the privacy mechanism.

%% file: paper/framework.tex
\textbf{Notation:} Let $[d]$ denote the set $\{1,2,\ldots,d\}$. The $\ell_0$-norm of a vector is the number of non-zero entries in the parameter. For any $S \subset [d]$, define $v_{S} = [v_{1,S}, \ldots, v_{d,S}]$, where, $v_{i,S} = v_{i}\mathbf{1}\left( i \in S\right)$. 
For a matrix $M \in \mR^{d \times d}$, let $M_{S}$, denote the set of columns of $M_{S}$ that correspond to indices in $S$. The support of $v \in \mR^{d}$ denotes the non-zero indices of $v$. 
We consider a linear contextual bandit problem with $K$ arms, where the (unknown) underlying vector $\theta \in \mathbb{R}^d$ that parameterizes rewards is high-dimensional and sparse -- often called the LASSO bandit setting. We consider a time horizon $T$, where for each time $t \in [T]$, the algorithm is given an (exogenous) context vector $\cX_{t} = \{X_{t,k}\}_{k \in [K]}$, where each $X_{t,k} \in \mathbb{R}^d$. The $\cX_{t}$ are drawn i.i.d. from an unknown distribution $\cD$. After observing $\cX_{t}$, the algorithm selects an arm $k_{t} \in [K]$ and observes a random reward $r_t$ given by $r_{t} = \langle X_{t,k_{t}}, \theta \rangle + \eta_{t}$, where $\eta_{t}$ is zero-mean bounded noise with variance $\sigma^{2}$. Further, $\Vert \theta \Vert_{2} \le C_{\theta}$ and $ \vert r_t \vert \le C_r$ for some positive constants $C_{\theta}$ and $C_r$.
Let $\cH_t$ denote the tupple of random variables generated by the past contexts, arm pulls, and observed rewards and $\cF_{t}$ denote the corresponding filtration. We assume that the parameter vector $\theta \in \mR^{d}$ is $s$-sparse, i.e., its support $S = \{i \in [d]: \theta_{i} \neq 0 \}$ has cardinality $s$, which is known to the decision maker. The algorithm's goal is to minimize cumulative expected regret, defined as $R_T = \textstyle\sum_{t=1}^{T} \mE[ \max_{k \in [K]} \langle X_{t,k} - X_{t,k_t} , \theta \rangle].$

\subsection{Model Assumptions}
We now describe the assumptions necessary to establish our results. Our first assumption implies that both the contexts and the parameters remain bounded, and is crucial is establishing both privacy and regret guarantees. 
\begin{assumption}[Boundedness]
\label{assmpt:bounded}
For any subset $S' \subseteq [d]$ such that $\vert S' \vert \le O(s)$, the $\ell_{2}$-norm of the context vector is bounded. That is, given constants $C_1$ and $\phi$ 
for all $t \in [T], \ k \in [K],\ \ X_{t,k} \in \cX_t$, and $S' \subset [d], \ \text{such that}\  \vert S' \vert \le s + \frac{4 C_{1}\sqrt{s}}{\phi^{2}}$, there exists $C_{1} >0$ such that $\Vert \cX_{S'} \Vert_{2} \le C_{1}$. 
\end{assumption}
Our strongest assumption is that the observed reward is bounded. This is necessary for us to ensure the privacy of the observations since the noise added to preserve privacy depends on the magnitude of this reward function, see Section~\ref{sec:sparsethresholdLASSO}. The following compatibility condition is standard in high-dimensional statistics literature~\cite{buhlmann2011statistics}. It implies that the $\ell_1$-regularized estimate converges to the true parameter both in value (consistency) and in support as the number of samples increases. The assumption holds for popular covariate distributions such as the multivariate Gaussian and the uniform distribution. When the eigenvalue of the sample covariance matrix is lower bounded by a constant, then the compatibility constant is lower bounded by the same constant.

\begin{definition}[Compatibility]
For a matrix $M \in \mR^{d \times d}$ and a set $S \subseteq [d]$, the compatibility constant $\phi(M,S)$ is defined as $\phi^{2}(M,S) = \min_{x: \Vert x_{S} \Vert_{1} \neq 0} \left\{ \frac{x_{S}^{\top}Mx_{S}}{\Vert x_{S} \Vert_{1}^{2}} :   \Vert x_{S^{c}}\Vert_{1}\le 3 \Vert x_{S} \Vert_{1}\right\}$.
\end{definition}

\begin{assumption}[Compatibility Condition]
\label{assmpt:compatibility}
The gram matrix of the action set $\Sigma:= \frac{1}{K} \sum_{k=1}^{K} \mE_{\cX \sim \cD} [X_{t,k}X_{t,k}^{\top}]$ satisfies $ \phi^{2}(\Sigma, S) \ge \phi^{2}$ for some $\phi >0$. 
\end{assumption}
 
Our next two assumptions were introduced in~\cite{oh2021sparsity}. The symmetry requirement of Assumption~\ref{assmpt:symmetry} holds for popular sampling distributions such as the multivariate Gaussian, uniform distribution on the sphere, and other isotropic distributions. Assumption~\ref{assmpt:balanced} implies that the contexts are diverse enough around the parameter $\theta$ to enable learning.

\begin{assumption}[Relaxed Symmetry]
\label{assmpt:symmetry}
For a distribution $\cD$ of $\cX$, there exists a constant $r_\cX > 1$ such that for all $\cX \in \mR^{K \times d}$, $\cD(\cX) > 0$  and $\frac{\cD(\cX)}{\cD(-\cX)} \le C_s$. 
\end{assumption}

\begin{assumption}[Balanced Covariates]
\label{assmpt:balanced}
For any permutation $\pi$ of $[K]$, for any integer $\{2,3,\ldots,K-1\}$ and a fixed $\theta$, there exists a constant $C_b > 1$ such that: 
\begin{eqnarray*}
&& C_b \mE_{\cX \sim \cD} [ (\cX_{\pi(1)}\cX_{\pi(1)}^\top+\cX_{\pi(K)}\cX_{\pi(K)}^\top)\mathbf{1}( \langle \cX_{\pi(1)} ,\theta\rangle, \langle \cX_{\pi(2)},\theta \rangle, \ldots, \langle \cX_{\pi(K)} , \theta \rangle)] \\
&\ge& \mE_{\cX \sim \cD}[\cX_{\pi(K)}\cX_{\pi(K)}^\top\mathbf{1}( \langle \cX_{\pi(1)} ,\theta\rangle, \langle \cX_{\pi(2)},\theta \rangle, \ldots, \langle \cX_{\pi(K)} , \theta \rangle )].
\end{eqnarray*}
\end{assumption}

Assumption~\ref{assmpt:sparse-pd} is an extension of a well-established assumption in low-dimensional bandit literature~\cite{degenne2020gamification,hao2020adaptive}. In low-dimensional regimes, this assumption implies that the set of context vectors spans the space $\mR^d$. In high-dimensional regimes, this assumption is used to show that the error in the least-squares estimate is $O(s)$.

\begin{assumption}[Sparse Positive Definiteness]
\label{assmpt:sparse-pd}
For each $B \subseteq [d]$, let $\Sigma_B = \frac{1}{K}\sum_{k=1}^K \mE_{\cX \sim \cD} X_{B}X_{B}^\top $. There exists a positive constant $C_p$ such that for all $B \subseteq [d],\ \min_{\Vert y \Vert_2=1} y^\top \Sigma_B y \ge C_p$. 
\end{assumption}
Define $\underbar{s} = \left(1+ \frac{4C_r C_\cX}{\phi^2}\right)$ and $\bar{s} = \left(1+ \frac{4C_r C_\cX\sqrt{s_0}}{\phi^2}\right)$ and for $ t \in [2^\ell, 2^\ell-1], s_{1,\ell}:= s +\frac{16sC_rC_{\cX}\lambda_\ell}{\phi^2[4\lambda_\ell\Gamma \sqrt{s (1+\frac{4 C_r C_\cX}{\phi^2})}+\alpha]}$ to be used in subsequent results.

\subsection{Privacy Preliminaries}
We now formalize the privacy definition for the bandit problem. We begin with the standard notion of differential privacy in streaming settings. In \emph{streaming settings}, data points arrive one at a time, and the algorithm must make a decision after each timestep. This is in contrast to static settings, where the entire database is known upfront and the algorithm only needs to make a single decision based on the full database. Two streams $\cS=\{(\cX_{s}, r_{s})\}_{s=1}^{T}$ and $\cS'=\{(\cX'_{s}, r'_{s})\}_{s=1}^{T}$ are said to be $t$-neighboring if for all $t' \neq t$, it holds that $(\cX_{t'}, r_{t'}) = (\cX'_{t'}, r'_{t'})$. Let $\cH = \mathbb{R}^{d \times k} \times \mathbb{R}, \ \cS \in \cH^{T}$. 

\begin{definition}[Differential Privacy for Streams \cite{DMNS06,CPW16}]
\label{defn:eps-delta-dp}
A streaming algorithm $\cA: \cH^T \to  [K]^{T} $ is  $(\varepsilon,\delta)$-differentially private for any pair of $t$-neighboring streams $\cS,\cS'$ and any $\cT \subseteq  [K] \times \ldots \times [K]$,
\[\Pr[ \cA(\cS) \in \cT] \le \exp(\varepsilon) \Pr[ \cA(\cS') \in \cT]+\delta.\]
When $\delta=0$, we may refer to $\cA$ as \emph{$\varepsilon$-differentially private}.
\end{definition}

Differential privacy ensures that the output of the mechanism is insensitive to changing a single element of the input.  However, in our contextual bandit setting using the standard notion of differential privacy in streaming settings implies that the choice of arm at time $t$ should be nearly independent of the context $\cX_t$, which is suboptimal in bandit settings. We instead use \emph{joint differential privacy} (JDP), first introduced by \cite{KPRU14} in the context of algorithmic mechanism design, and later extended to the online bandit setting by \cite{shariff2018differentially}. This slight relaxation of differential privacy allows the $t$-th component of the output (i.e., $k_t$) to depend arbitrarily on the $t$-th component of the input (i.e., $\cX_t$), while ensuring differential privacy concerning the joint distribution of all other components of the output. 

\begin{definition}[Joint Differential Privacy for Streams \cite{KPRU14, shariff2018differentially}]
A streaming algorithm $\cA: \cH^T \rightarrow [K]^T$ is said to be $(\varepsilon, \delta)$-jointly differential private, for any $t$-neighboring streams $\cS,\cS'$ and any $\cT \subseteq [K]^{T-1}$, 
\[ 
\Pr[ \cA(\cS)_{\neq t} \in \cT] \le \exp(\varepsilon) \Pr[\cA(\cS')_{\neq t}  \in \cT ] + \delta,\]
where $\cA(\cS)_{\neq t}$ denotes all portions of the algorithm's output except at time $t$.
\end{definition}

The primary algorithmic way of achieving joint differential privacy is via the \emph{Billboard Lemma} below (Lemma \ref{lem:billboard}). It says that to achieve joint differential privacy, it is sufficient to publish a global statistic under differential privacy (e.g., on a metaphorical billboard), and then allow each individual to compute her portion of the output using only this global statistic and knowledge of her input. In our bandit setting, we will compute a differentially private estimate $\hat{\theta}^{\text{priv}}_t$ of the unknown parameter vector $\theta$; from this estimate, the algorithm can compute an estimate of the best arm $k_t$ and receive the reward $r_t$ using only this estimate and knowledge of the context $\cX_t$.
Further, we would like to point out that we are preserving the privacy of the entire dataset as opposed to the observed rewards as in~\citep{hanna2022differentially}.

\begin{lemma}[Billboard Lemma \cite{HHRRW14,RR14}]
\label{lem:billboard}
Suppose $\cA : \cU \to \cV$ is $(\varepsilon,\delta)$-differentially private. Consider any set of functions $f_t: \cU_t \times \cV \to \cV_0$, where $\cU_t$ is the portion of the database containing the $t$-th component of the input data. Then the composition $\{f_t(\Pi_t(\cS),\cA(\cS))\}_{t\in[T]}$ is $(\varepsilon,\delta)$-jointly differentially private, where $\Pi_t(\cS): \cU \to \cU_t$ is the projection onto the $t$-th input data coordinate.
\end{lemma}

Differential privacy \emph{composes adaptively}, meaning that the privacy guarantees degrade gracefully as additional private computations are run on the same data (Proposition \ref{lem:advanced-composition}). It also satisfies \emph{post-processing}, meaning that any downstream computation performed on the output of a differentially private algorithm will retain the same privacy guarantee (Proposition \ref{prop:post-processing}).

\begin{proposition}[Advanced Composition \cite{DRV10}]
\label{lem:advanced-composition}
For any $\varepsilon > 0$, $\delta \geq 0$, and $\delta' > 0$, the $T$-fold adaptive composition of $(\varepsilon,\delta)$-differentially private mechanisms satisfies $(\varepsilon',k\delta+\delta')$-differential privacy for $\varepsilon' = \sqrt{2k\ln(\frac{1}{\delta})\varepsilon} + k\varepsilon(\exp(\varepsilon)-1)$.
\end{proposition}

\begin{proposition}[Post Processing \cite{DMNS06}]
\label{prop:post-processing}
Let $\cA: \cU \to \cV$ be $(\varepsilon,\delta)$-differentially private, and let $f: \cV \to \cV'$ be any randomized mapping. Then, $f \circ \cA: $ is $(\varepsilon, \delta)$-differentially private.
\end{proposition}

We will use two well-known tools for designing our algorithms: \textit{sparse vector technique} and \textit{tree-based aggregation protocol} (TBAP). The \textit{Sparse Vector Technique} is used
to privately count the number of entries above a predetermined threshold.  It requires the knowledge of the count of the number of non-zero entries in the underlying vector and a threshold to construct the support of the parameter vector. To reconstruct the support of the sparse vector, the threshold is perturbed with appropriate noise and then used for filtering the observations. 
We will use a \emph{tree based aggregation protocol} (TBAP), introduced by~\cite{chan2011private,dwork2010differential} for binary data and extended by \cite{ST13} to real-valued data, to preserve the privacy of data arriving in an online stream. At each time $t$, the algorithm will produce a differentially private estimate of the partial sums of all data seen up until time $t$. \cite{mishra2015nearly} showed how to design a binary tree-based aggregation which ensures $\cO\left( \frac{\log^{1.5}T}{\varepsilon} \right)$ error and $(\varepsilon,0)$-differential privacy. Note that this dependence on $T$ is substantially better than what would arise under advanced composition. We will use a tree-based aggregation protocol to carry out online private linear regression. TBAP is used to privately compute the $\ell_2$-regression estimate of the true parameter $\theta$ by recording intermediate estimates as nodes of the tree. Since the $\ell_2$ estimate can be decomposed as a sum, such a structure provides a significant advantage as the estimate at time $t$, can be computed in $\log t$ time.

%% file: paper/threshold-lasso.tex
In this section, we present our main algorithm for privately learning LASSO bandit, PrivateThresholdLASSO (Algorithm~\ref{algo:private-lasso}). 
PrivateThresholdLASSO is an episodic hard-thresholding-based algorithm that privately maintains an estimate of the support of the parameter vector $\theta$. At each time $t$, the algorithm observes the set of context vectors $\cX_t$ to decide the arm to be played. After observing the context $\cX_t$, first, it privately estimates the support of $\theta$ (presented formally in Section \ref{s.sparseset}) if we are entering a new episode (Line~\ref{algo-step:sparse-estimation} of Algorithm~\ref{algo:private-lasso}). The sparse estimation step takes as an input the previous history of the algorithm and various pre-known parameters such as the privacy budget to compute private support, $S_t$ of the parameter $\theta$. This is described in Section~\ref{s.sparseset}. This estimate of the support remains fixed throughout a given episode. Then, we estimate $\theta_t$ to be used for taking an action at time $t$. This is done by performing $\ell_2$-regression on the entire history of observations up to time $t$ but restricting the non-zero estimates to indices corresponding to those in $S_t$ (Line~\ref{algo-step:tbap} of Algorithm~\ref{algo:sparse-estimation}). This leads to a sparse estimate of $\theta_t$. Finally, the action played is the best action assuming $\hat{\theta}_t$ to be the true parameter.
\label{sec:sparsethresholdLASSO}
\begin{algorithm}[tbh]
\caption{PrivateThresholdedLASSO}
\label{algo:private-lasso}
\begin{algorithmic}[1]
\STATE{Input: $\varepsilon, \delta, \lambda_{0}, \{\cX_s\}_{s=1}^{t}$, Output: $\{k_s\}_{s=1}^{t}$} 
\STATE{Initialize: $\delta_{1},\delta_{2} \leftarrow \frac{\delta}{2\lceil \log T \rceil}, \frac{\delta}{2\lceil \log T \rceil}, \ \varepsilon_{1}, \varepsilon_{2} \leftarrow \frac{\varepsilon}{2\lceil \log T \rceil \ln(\frac{1}{\delta_{2}})},\ \frac{\varepsilon}{2\lceil \log T \rceil \ln(\frac{1}{\delta_{2}})}$}
\STATE{$L \leftarrow \left\{ t \in [T]: t= 2^{k-1}, \; k=1, \ldots, \lfloor \log_{2} T \rfloor +1 \right\}$}
\FOR{$t \in [T]$}
\STATE{Observe context $\cX_{t} = \{X_{t,k}, k \in [K]\}$}
\IF{$t \in L$}
\STATE{$\lambda_{t} \leftarrow \lambda_{0} \sqrt{\frac{2\log t\log d}{t}}$}
\STATE{$S_{t} \leftarrow \text{SparseEstimation}$ (See Algorithm~\ref{algo:sparse-estimation})}\label{algo-step:sparse-estimation}
\ELSE
\STATE{$S_{t} \leftarrow S_{t-1}$}
\ENDIF
\STATE{$\hat{\theta}^{\text{priv}}_t \leftarrow \text{TBAP}$}  
\label{algo-step:tbap}
\STATE{Play arm $k_{t} \leftarrow \arg\max_{k \in [K]} X_{t,k}^{\top}\hat{\theta}^{\text{priv}}_{t}$}\label{algo-step:play-arm}
\STATE{Receive reward $r_t = \langle X_{t,k_{t}}, \theta \rangle + \eta_{t}$}
\STATE{Update design matrix $Z = [X_{1,k_1}, X_{2,k_2}, \ldots, X_{t,k_t}]$ and regressor $Y = [r_1, r_2, \ldots, r_t]$}
\ENDFOR
\end{algorithmic}
\end{algorithm}

For the sparse $\ell_2$-regression step in line \ref{algo-step:tbap}, we propose a non-trivial adaptation of the tree-based aggregation protocol to reduce the noise that would result from privacy composition. Given the historical (context, reward) pairs $\{(X_{s}, r_s)\}_{s \le t}$, our goal is to construct a private estimate of the parameter $\theta$ using $\ell_2$-regression, given by $\hat{\theta}^{\text{priv}}_t := \arg\min_{\Vert \theta \Vert \le C_\theta} \Vert R - X_{S_t}^\top\theta\Vert^2$. We use TBAP to maintain private copies of the gram matrix $X_{s,k_s}X_{s,k_s}^\top$ and regressor $X_s r_s$. Instead of maintaining two binary trees, one for the gram matrix and another for the regressor, we maintain a single binary tree by constructing a single matrix $G_t \in \mR^{(d+1) \times (d+1)}$ as in~\cite{sheffet2015private} that stores all of this information. Let $B_t = \left[X_{1:t} , r_{1:t}\right] \in \mR^{t \times (d+1)}$ and $G_t = B_tB_t^\top$ and $G_{t+1} = G_{t} + [X_{t,k_t}^\top, r_t] 
\begin{bmatrix}
X_{t,k_t} \\
r_t
\end{bmatrix}$.
We will preserve the privacy of each node of the binary tree $\cB$ using a Wishart Mechanism, $\cW_{d+1}(\Sigma,k)$, which is a sample $k$-independent samples from $(d+1)$-dimensional Gaussian Distributions, $\cN(0,\Sigma)$ and computes their Gram Matrix. 
To ensure $\left(\varepsilon,\delta \right)$-DP, every node in the binary tree needs to preserve $\left(\frac{\varepsilon}{\sqrt{8\log T \ln(\frac{2}{\delta})}}, \frac{\delta}{2\log T} \right)$. This can be done by sampling a matrix from $\cW_{(d+1)}\left(\tilde{s}I_{(d+1) \times (d+1)}, k\right)$ with $k = \lceil d\varepsilon^{-2} \ln\left(\frac{8d}{\delta}\right)\left(\frac{2}{\delta}\right)\rceil $. 
Alternatively, one can use the AnalyseGauss mechanism~\citep{dwork2014analyze} to preserve privacy within each node. However, it is not guaranteed to produce a positive semi-definite matrix.A naive repeated application of these steps at \emph{every} time $t\in[T]$ would result in error $\Omega(\sqrt{T}/\varepsilon)$ by Advanced Composition (Lemma \ref{lem:advanced-composition}), which would be unacceptable for large $T$. Instead, the algorithm performs the sparsity estimation only periodically, using an epoch-based schedule at time $t$ in the exponentially-space update set:  
$L = \{ t \in [T]: t= 2^{k-1}, \; k=1, \ldots, \lfloor \log_{2} T \rfloor +1 \}$. For times $t$ in the set $L$, we update the private estimate of the support $S_{t}$. The estimated support is used to estimate $\theta$ at all times until the next update. Since only $\cO(\log T)$ updates will occur, the error due to privacy composition is $\cO(\log T/\varepsilon)$. We formally present the utility and the privacy guarantee of this algorithm in Section~\ref{sec:privacy} and~\ref{sec:regret}.

\subsection{SparseEstimation}
\label{s.sparseset}
In the SparseEstimation subroutine (Algorithm~\ref{algo:sparse-estimation}), we construct a sparse private estimate of the support of parameter $\theta$ based on  $\cH_t$ based on thresholding-based procedure (such as that used in~\cite{thlasso}). The algorithm begins by initializing pre-tuned parameters in Line~\ref{algo-step:initialize} and constructing an initial sparse estimate $\hat{\theta}$ using $\ell_{1}$-regularization, and then uses two-step thresholding-based procedure to further refine and identify the support. We use a two-step support estimation procedure: the first support estimate $S_0$ is constructed non-privately while the subsequent support, $S_1$ is constructed using a privatization mechanism, Sparse Vector Technique. The first $\ell_1$ estimate (Line~\ref{algo-step:lasso-estimate} in Algorithm~\ref{algo:sparse-estimation}) is performed using LASSO~\cite{tibshirani1996regression} with design matrix: $ Z_t = [X^\top_{1,k_1}, X^\top_{2,k_2} \ldots, X^\top_{t,k_t}] \in \mR^{t \times d} $ and regressors: $Y_t = [r_1, r_2, \ldots, r_t ]^\top$. The LASSO solution is the minimizer of the following convex program: 
\[
\hat{\theta}_t = \arg\min_{\theta: \Vert \theta \Vert_2 \le C_{\theta} } \Vert  Y_t - Z_t^\top \theta \Vert_2^{2} + \lambda_t \Vert \theta \Vert_2 
\]

\begin{algorithm}[tbh]
\caption{SparseEstimation}
\label{algo:sparse-estimation}
\begin{algorithmic}[1]
\STATE{Initialize: 
\scalebox{0.8}{$\varepsilon' =  \varepsilon_1\left(8\bar{s}\ln(\frac{1}{\delta_1}))\right)^{-1/2},
\xi \leftarrow \frac{\left(32  \ln(\frac{1}{\delta_1})\right)^{1/2}}{\varepsilon'}, \Gamma
= \bar{s}^{-1/2}\left[\frac{\xi\log (\frac{1-\delta}{\varepsilon})-C_\theta}{\sqrt{\log d \log T}}-1\right]$
}}
\label{algo-step:initialize}
\label{algo-step:startsv}
\STATE{$\hat{\theta}_{t} \leftarrow \Vert Y_t -  Z_t\theta \Vert^2_{2} + \lambda_{t} \Vert \theta \Vert_{1}$}
\label{algo-step:lasso-estimate}
\STATE{$S_{0} \leftarrow \{i \in [d]: \vert \hat{\theta}^{(i)}_{t} \vert > 4\lambda_{t} \}$}
\label{algo-step:construct-s0}
\FOR{$i \in S_{0}$}
\label{algo-step:startsv}
\STATE{$\tilde{\lambda}_{t} \leftarrow 4\lambda_{t}\Gamma\sqrt{\bar{s}} +\zeta_i, \ \zeta_i \sim \text{Lap}(\xi)$}\label{algo-step:laplace-threshold}
\STATE{$\tilde{\theta}^{(i)}_t \leftarrow \hat{\theta}^{(i)}_{t} + \nu_{i},\ \nu_{i} \sim \text{Lap}(2\xi)$}
\IF{$ \tilde{\theta}^{(i)}_t > \tilde{\lambda}_{t}$}
\label{algo-step:theta-noise}
\STATE{$S_{1} \leftarrow \{i\} \cup S_{1}$}
\STATE{$c = c+1$}
\ENDIF
\IF{$c \ge s_0+\sqrt{\bar{s}}$}
\STATE{Return $S_{1}$}
\ENDIF
\ENDFOR 
\label{algo-step:endsv}
\end{algorithmic}
\end{algorithm}

Line~\ref{algo-step:construct-s0} in Algorithm~\ref{algo:sparse-estimation} is a non-private thresholding-based estimation of support, $S_0$. $S_0$ uses a much lower (non-noisy) threshold than $S_1$, so as to not screen out any of the non-zero coordinates in this step. Our motivation for using a two-step thresholding is two-fold: (i) the second step is the primary privacy preserving step in our setting and  (ii) such procedures have demonstrated significant performance improvements in the accuracy of estimation in non-private setting~\citep{belloni2013least} and in reducing the number of false-positives albeit at the cost of increase in estimation bias. The thresholding step (Lines \ref{algo-step:startsv}--\ref{algo-step:endsv}) is based on the Sparse Vector Technique~\citep{DNRRV09,dwork2010differential,dwork2014algorithmic}. These steps sequentially check all coordinates of the initial estimate $\hat{\theta}_t$ to determine those which are non-zero via thresholding. As in the Sparse Vector Technique, privacy is ensured by comparing a noisy version of the coordinate value with a noisy version of the threshold. Those coordinates identified to be above the noisy threshold are added to the set $S_1$ of coordinates estimated to be non-zero. A counter is maintained in to ensure that not too many coordinates are identified by the algorithm as being non-zero. This counter benefits both privacy -- by ensuring that privacy composition occurs over only a small number of runs of AboveNoisyThreshold -- and sparsity -- by ensuring that only a small number of non-zero coordinates are returned.  Lemma~\ref{lem:support-recovery} in Section \ref{sec:privacy} analyses the effect of the private thresholding step on the recovered support of $\theta$ and its size.  The returned support $S_1$ is then used to construct an estimate of the parameter $\theta$ using online private linear regression in Line~\ref{algo-step:tbap} of Algorithm~\ref{algo:private-lasso}. There are two key implications of using such a two-step thresholding procedure: the size of estimated support, $\vert S_1 \vert$ is larger than in the non-private counterpart and (ii) it is possible when used on two neighbouring sequences the additional thresholding step reveals information about the difference in support between $S_0$ and $S_1$. To account for this scenario, we increase the threshold so that the probability of such an event happening is accounted for in $\delta$, leading to $\delta >0$.

%% file: paper/privacy.tex
In this section, we present the privacy guarantees of PrivateLASSO (Theorem~\ref{thm:private-lasso}), which is our first main result. At a high level, the privacy guarantees of the estimates $\hat{\theta}_t^{\text{priv}}$ come from the composition of two major components of Algorithm~\ref{algo:private-lasso}: SparseEstimation (Algorithm~\ref{algo:sparse-estimation}) and Tree-based aggregation. We will separately analyze the privacy of these two components and then combine their guarantees to establish Theorem~\ref{thm:private-lasso}.

\begin{lemma}[Privacy of SparseEstimation, Algorithm~\ref{algo:sparse-estimation}]
\label{lem:privacy-sparse-estimation}
SparseEstimation is $(\varepsilon_1,\delta_1)$-differentially private.
\end{lemma}

The SparseEstimation algorithm is based in large part on $(\varepsilon,0)$-differentially private AboveNoisyThreshold where the coordinates of $\hat{\theta}_t$ play the role of queries, so its privacy guarantees follows from AboveNoisyThreshold~\cite{DNRRV09,dwork2010differential,dwork2014algorithmic}. However, some algorithmic modifications require careful privacy analysis. In particular, the algorithm has a non-private thresholding step to select a subset $S_0$ of coordinates in $\hat{\theta}_t$ that are candidates for being non-zero. Only coordinates that are above the non-noisy threshold for $S_0$ will be considered in the AboveNoisyThreshold-portion of the algorithm. Naturally, this creates the possibility of information leakage. In particular, consider a pair of neighboring databases $\cS,\cS'$ where the value of a coordinate $\hat{\theta}^{(i)}_t$ was above the $S_0$ threshold in $\cS$, and below the threshold in $\cS'$. However, because of the large gap between the threshold values for $S_0$ and $S_1$, we show that $\hat{\theta}^{(i)}_t$ when observations come from $\cS$ must be far from the threshold for $S_1$ and the probability of sampling a noise term $\zeta$ that is large enough to cause it to be included in $S_1$ is exponentially small. We account for this in the $\delta$ privacy parameter; a similar analysis was used for a non-private thresholding step in \cite{zhang2020paprika}.  While Lemma~\ref{lem:privacy-sparse-estimation} ensures that the recovered support of $\theta$ is private, the estimate of $\theta$ further relies on tree-based aggregation protocol. TBAP uses a Wishart mechanism to preserve the privacy of entries in a tree node. The privacy guarantee of this procedure then directly follows by adaptively composing the privacy guarantee for TBAP and the Wishart Mechanism. The proof of Theorem~\ref{thm:private-lasso} combines Lemma~\ref{lem:privacy-sparse-estimation} with known privacy guarantees for TBAP~\cite{chan2011private,dwork2010differential,mishra2015nearly}. 

\begin{theorem}[Privacy Guarantees for Algorithm~\ref{algo:private-lasso}]
\label{thm:private-lasso}
In PrivateLASSO, the sequence of estimates $\{\hat{\theta}^{\text{priv}}_t\}_{t \in [T]}$ is $(\varepsilon,\delta)$-differentially private, and the sequence of arm pulls $\{k_t,\}_{t \in [T]}$ are $(\varepsilon,\delta)$-jointly differentially private. 
\end{theorem}

Choosing $\varepsilon_1, \varepsilon_2, \delta_1, \delta_2$ as stated in Algorithm~\ref{algo:private-lasso} along with basic composition across these two subroutines will ensure that the sequence of vectors $\{\hat{\theta}^{\text{priv}}_t\}_{t\in [T]}$ produced by Algorithm~\ref{algo:private-lasso} satisfy $(\varepsilon,\delta)$-differential privacy. Using the Billboard Lemma (Lemma~\ref{lem:billboard}) we can then conclude that the sequence of contexts, arm pulls and rewards $\{(k_t,r_t)\}_{t \in [T]}$ are $(\varepsilon,\delta)$-jointly differentially private.

%% file: paper/regret.tex
In this section, we show that PrivateLASSO (Algorithm~\ref{algo:private-lasso}) has $\cO(\frac{s^{3/2}_0\log^{3}T}{\varepsilon})$ regret under the commonly-assumed \emph{margin condition} (Theorem \ref{thm:regret-margin}) and $\cO(\frac{s^{3/2}_0\sqrt{T}\log^2 T}{\varepsilon})$ without this assumption (Theorem \ref{thm:regret-without-margin}). In Section~\ref{s.accesttheta}, we study the impact of noise added for privacy on the accuracy of our estimate $\hat{\theta}^{\text{priv}}_t$. Then in Sections \ref{s.regretmargin} and \ref{s.withoutmargin}, we express the per-episode regret in terms of this estimation error and present formal statements of Theorems \ref{thm:regret-margin} and \ref{thm:regret-without-margin}, respectively. 
\subsection{Utility Gaurantees}\label{s.accesttheta}
We begin by bounding the error our estimation of $\theta$, orignating from estimation as well as noise added to preserve privacy.  In Lemma~\ref{lem:sparse-estimation-accuracy}, we show that $S_1$, our private estimate of the support of the parameter $\theta$ from the SparseEstimation subroutine (Algorithm \ref{algo:sparse-estimation}) does not suffer from too many false positives or false negatives. This allows us to (approximately) accurately recover the true support of $\theta$ (Lemma~\ref{lem:support-recovery}). Finally, we use these results, along with bounds on the error due to using Tree-based aggregation protocol, to bound the error of our estimate $\hat{\theta}_t^{\text{priv}}$ (Lemma~\ref{lem:estimation-error}). We first introduce our accuracy notion for SparseEstimation (Algorithm~\ref{algo:sparse-estimation}). Error in this algorithm can come from misclassifying the coordinates of $\hat{\theta}_t$ as being above or below the threshold $\tilde{\lambda}_t$. Definition \ref{defn:alpha-beta-accuracy} requires that values determined to be above (resp. below) the threshold are not more than $\alpha$ below (resp. above) the threshold.

\begin{definition}[Estimation Accuracy]
\label{defn:alpha-beta-accuracy}
An algorithm $\cA: \mathbb{R}^{\vert S_0 \vert} \to 2^{ S_0 }$ that takes in $|S_0|$ values, $\theta_1, \ldots, \theta_{\vert S_0 \vert}$, and outputs a subset $S \subseteq S_0$ is \emph{$(\alpha,\frac{1}{T})$-accurate} with respect to a threshold $\lambda$ if for all $i \in S, \; \theta_i \ge \lambda -\alpha$, and for all $i \notin S, \; \theta_i \le \lambda+\alpha$ with probability $1-\frac{1}{T}$.
\end{definition}

Lemma~\ref{lem:sparse-estimation-accuracy} largely follows from known accuracy bounds of AboveNoisyThreshold \cite{DNRRV09,chan2011private}, upon which Algorithm~\ref{algo:sparse-estimation}. 
\begin{lemma}[Accuracy of Algorithm~\ref{algo:sparse-estimation}]
\label{lem:sparse-estimation-accuracy}
SparseEstimation (Algorithm~\ref{algo:sparse-estimation}) is $(\alpha,\frac{1}{T})$ accurate for $\alpha = \left(\frac{128\bar{s}}{\varepsilon}\ln(\frac{1}{\delta})\right) \ln\left(2Ts^{3/2}\underbar{s}\bar{s}\right)$. 
\end{lemma}
Lemma~\ref{lem:sparse-estimation-accuracy} shows that accuracy of the thresholding step degrades inversely with $\varepsilon$ and increases only logarithmically in $T$. 
This result can now be used to both ensure that the true non-zero coordinates of $\theta$ are included in our estimate $\hat{\theta}_t^{\text{priv}}$ and to lower bound the size of its support. To this end, Lemma~\ref{lem:support-recovery} uses the accuracy result of Lemma~\ref{lem:sparse-estimation-accuracy} to ensure support recovery in the private setting inspired from the non-private setting~\cite{thlasso}. Fix episode $\ell \in [L]$ and define the event $\cE_{1,\ell} = \{t \in [2^\ell, 2^{\ell+1}-1]: \Vert \hat{\theta}^{\text{priv}}_t - \theta \Vert_1 \le \tfrac{4s \lambda_{t}}{\phi^2_{t}} \}$, which is the event that the estimation error in $\hat{\theta}^{\text{priv}}_t$ is no more than $\frac{4s \lambda_{t}}{\phi^2_{t}}$. 
\begin{lemma}[Support Recovery]
\label{lem:support-recovery}
For episode
$\ell \in [L]$ such that $4\lambda_{\ell}\left(\frac{4 C_s C_{\cX}s}{\phi^{2}}+\sqrt{(1+\frac{4r_\cX C_{\cX}}{\phi^{2}}})s \right) \le \theta_{min}$ and $\lambda_\ell \ge (\frac{128\bar{s}}{\varepsilon}\ln(\frac{2s^{3/2}\bar{s}\underbar{s}}{\beta})\ln(\frac{1}{\delta})$. 
If Algorithm~\ref{algo:sparse-estimation} is $(\alpha,\beta)$-accurate and under Assumptions~\ref{assmpt:bounded},~\ref{assmpt:compatibility},~\ref{assmpt:symmetry} and~\ref{assmpt:balanced} and event $\cE_{1,\ell}$, the support set $S_{\ell}$ estimated in episode $\ell$ in Algorithm \ref{algo:private-lasso} will satisfy for all $t \in [2^\ell, 2^{\ell+1}-1]$, 
\[\mP\left[ S \subset S_{\ell} \text{ and } \vert S_{\ell} \setminus S \vert \le 
s_{1,\ell}\right] \ge 1- \exp( \tfrac{-t \lambda^{2}_{\ell}}{32 \sigma^{2} C^{2}_{\cX}} + \log d) -\exp(\tfrac{-tC^{2}}{2})-\frac{1}{T}.
\]
\end{lemma}
Lemma~\ref{lem:support-recovery} is proved by bounding the $\ell_1$-error of the parameter estimates in the support $S_\ell$ using Theorem \ref{lem:sparse-estimation-accuracy} and the assumed bound on the empirical covariance matrix of the observed context vectors. (Assumption \ref{assmpt:compatibility}).Lemma~\ref{lem:support-recovery} establishes the price of privacy for accurate support recovery.In the non-private setting, the number of false positives is upper bounded by $\frac{4C_rC_{\cX}\sqrt{s}}{\phi^2}$ which is roughly the $\ell_1$ error in estimation of $\theta$ divided by the non-private threshold, $4\lambda_t$.  When constructing a private estimate, this threshold increases by a factor of $4\sqrt{s_0}\lambda_{\ell} / \left(4\lambda_\ell \Gamma\sqrt{s_0(1+\tfrac{4 r_\cX C_\cX}{\phi^2})}+\alpha\right)$ due to the additional noise injected to preserve privacy. The affect of privacy is reflected in $\alpha$. As we increase $\varepsilon$ (low privacy), $\alpha \rightarrow 0$ and we recover the results for non-private case. Further, as a result of this inflated threshold, the estimated private support contains a larger number of false positives. In light of Lemma~\ref{lem:support-recovery}, we define the event $\mathcal{E}_{2,\ell} := \{t \in [2^\ell, 2^{\ell+1}-1]: S \subset S_{t} \ \text{and} \ \vert S_{t} \vert \le s_{1,\ell} \}$ which holds with high-probability due to Lemma \ref{lem:support-recovery}. Finally, when the support of $\theta$ is estimated correctly, $\cG_{\ell} = \cE_{1,\ell} \cap \cE_{2,\ell}$. We will condition the remainder of our analysis on the event $\cG_{\ell}$. We now establish a bound on the estimation error in $\theta$ when using the private estimate $\hat{\theta}^{\text{priv}}_t$, $\Vert \theta - \hat{\theta}^{\text{priv}}_t \Vert$. The estimation error can be decomposed due to the contribution of noise and covariance matrix in Lemma~\ref{lem:estimation-error}. We show that the minimum eigenvalue of the gram matrix increases at a sufficient rate. This is essential to ensure that the algorithm performs sufficient exploration so that the private estimates $\hat{\theta}^{\text{priv}}_t$ converge sufficiently fast to the correct parameter $\theta$. Lemma~\ref{lem:estimation-error} establishes the error in an estimate due to TBAP along with estimation error due to $\ell_2$-regression to establish the desired bound. 
\begin{lemma}[Estimation Error]
\label{lem:estimation-error}
For $\ell \in [L]$ and $t \in [2^{\ell},2^{\ell+1}-1]$ for all $x, \lambda > 0$,
\[
\mP( \Vert \hat{\theta}^{\text{priv}}_{t} - \theta \Vert_{2}  > x \ \text{and} \ \lambda_{\min}(\Sigma_{S_t}) \ge \lambda \; \vert \; \mathcal{G}_{\ell} ) \le 2s_1\exp(\tfrac{-\lambda^2t x^2}{2(\sigma^2+\sigma^2_{\cB})C^2_\cX s_1}) + 2\exp(-3\sigma^2_{\cB}s_1). \]
\end{lemma}

In the following subsections, this bound will be used to derive bounds on the regret of Algorithm \ref{algo:private-lasso}, which will rely on accuracy of the estimator $\hat{\theta}^{\text{priv}}_{t}$ with respect to $\theta$.

\subsection{Regret bounds under the Margin Condition}\label{s.regretmargin}
We first bound the regret of Algorithm \ref{algo:private-lasso} under the \emph{margin condition}, which formalizes the difficulty of identifying the optimal arm given a set of contexts. It was introduced in the bandit literature by~\cite{goldenshluger2013linear} and has since become standard in classification problems~\cite{audibert2007fast,tsybakov2004optimal,bastani2020online,oh2021sparsity}. We note that it is strictly weaker than requiring the \emph{gap assumption}, which is popular in finite-armed bandits~\cite{abbasi2011improved} and has been utilized in prioer private-bandit literature such as~\citep{sheffet2015private}.
\begin{assumption}[Margin Condition]
\label{assmpt:margin}
There exists $C_{\kappa} > 0$ such that for all $\kappa > 0$, and for all $k \neq k', \ \mP_{\cX \sim \cD} [ \vert  \langle X_{k} - X_{k'}, \theta \rangle \vert  \le \kappa ] \le C_{\kappa}\kappa.$
\end{assumption}

The margin condition identifies the hardness in distinguishing two context vectors that whose expected rewards are close to each other, and $\kappa$ controls the hardness of identifying the optimal arm when these two context vectors are observed. Loosely speaking, it implies that the expected regret is of the same order as the as the observed regret. Therefore, it is suffices to obtain a tight bound on the regret for different sample paths. For a hard-instance (hardness implying difficulty in identifying the optimal-arm for a given set of contexts), this implies that a larger regret is incurred because the chances of playing the sub-optimal arm are larger. The margin condition then relates the sub-optimality of the context-arm pair to the error in estimating $\theta$ through Lemma~\ref{lem:estimation-error}.  Lemma~\ref{lem:per-episode-margin} gives regret guarantees for PrivateLASSO under the margin condition. The margin condition allows us to tightly control the regret under the observed set of contexts at time $t$, and relates the per-episode regret to the squared estimation error, which we show decreases with the number of episodes. The per-episode regret is given by: $r(\ell) = \sum_{s=2^{\ell}+1}^{2^{\ell+1}-1} \mE[ \max_{k \in [K]} \langle X_{k} - X_{k_s} , \theta \rangle ]$. 
\begin{lemma}[Episodic Regret with margin condition]
\label{lem:per-episode-margin}
Under Assumptions~\ref{assmpt:bounded}--\ref{assmpt:sparse-pd} and~\ref{assmpt:margin}, the regret of Algorithm~\ref{algo:private-lasso} in episode $\ell$ is given by:
\begin{equation}
\label{eqn:per-episode-margin}
r(\ell) \le \frac{1048(1+\log 4s_{1,\ell})^{3/2}}{C_p^{2}2^{\ell}}(\sigma^2+\sigma^2_{\cB}) (K-1) C^4_{\cX}C^2_r C^2_b s_{1,\ell} + 2C_{\theta}C_{\cX} (K-1) \mP(\overline{\cG}_{\ell})
\end{equation}
\end{lemma}
Lemma~\ref{lem:per-episode-margin} is proved using a standard peeling argument while incorporating the effect of privacy. We break down the regret due to observed $\cX_t$ and using a private-estimator $\hat{\theta}^{\text{priv}}_t$. The margin condition, helps in simplifying the per-episode regret by discretizing the possible regret region in terms the sub-optimality of the context-arm pair. Finally, the error over all discretized regions is added to obtain the per-episode regret. The first term on the right hand side of~\eqref{eqn:per-episode-margin} is the contribution to regret when the good event holds (i.e, the parameters concentrate and the support set contains the true set with high probability) and the second term is the contribution due to the bad event. In Theorem~\ref{thm:regret-margin} we use the cumulative sum of the per-episode regret bound to establish the regret guarantee across all episodes. To obtain regret over all the episodes, in addition to bounding the per-episode regret under the good event $\mathcal{G}_\ell$ from Section \ref{s.accesttheta}, we show that the probability of the bad event, $\overline{\cG}_\ell$ remains small. Adding up the regret over different episodes leads to us to the final regret bond. 
\begin{theorem}[Regret with the margin condition]
\label{thm:regret-margin}
Under Assumptions~\ref{assmpt:bounded}--\ref{assmpt:sparse-pd} and~\ref{assmpt:margin}, the regret of Algorithm~\ref{algo:private-lasso} is given by: 
\begin{align*}
\cR(T) &\le  \frac{1048(1+\log 4s_1)^{3/2}}{C_p^{2}}(\sigma^2+\sigma^2_{\cB}) (K-1) s^4_{\cX}C^2_r C^2_\cX s_1 \log T \\
&\quad + 2(K-1)C_\theta C_\cX [64d(\sigma^2+\sigma^2_{\cB})C^2_\cX s^2_1\tfrac{1}{\lambda^2_0}\log T + \frac{2}{C^2_0} \log T] \\
&= \cO(\tfrac{s^{3/2}_0\log^3 T K \left(\sigma^{2}+\sigma^{2}_{\cB}\right) \log\frac{\log T}{\delta}}{\varepsilon})
\end{align*}
\end{theorem}
Theorem~\ref{thm:regret-margin} implies that the price of privacy for cumulative regret is $\cO(\frac{1}{\varepsilon})$, implying that higher privacy regimes incur larger regret. Furthermore, the regret scales linearly with the amount of noise added during the TBAP protocol which depends on the upper bound on the observed reward as well as the size of context vectors. The scaling concerning the time-horizon, number of arms, and sparsity parameters remains identical to the non-private setting.

\subsection{Regret bounds without the Margin condition}\label{s.withoutmargin}
We now extend our results without assuming the margin condition. While we still use a peeling argument, in the absence of the margin condition, the per-episode regret depends linearly on the estimation error. As a result, the per-episode regret decreases inversely with the square root of the episode length instead of being dependent on the inverse leading to a larger regret guarantee. 
\begin{lemma}[Episodic Regret without margin condition]
\label{lem:per-episode-without-margin}
Under Assumptions~\ref{assmpt:bounded}--\ref{assmpt:sparse-pd}, the per-episode regret of Algorithm~\ref{algo:private-lasso} is given by: 
\[
r(\ell) \le \frac{36\sqrt{1+\log 4s_1}}{C_p^{2}}(\sigma^2+\sigma^2_\cB) 
(K-1)C^2_{1} C^2_r C^2_\cX\frac{1}{\rho^2} \sqrt{2\left( s_0 + \tfrac{4 r_\cX C_\cX}{\phi^2_0} \right)\frac{1}{2^\ell}} + 2C_\theta C_{\cX} (K-1) \mP(\overline{\cG}_\ell).
\]
\end{lemma}
To get to total regret in Theorem~\ref{thm:regret-without-margin}, we sum over the regret incurred in every episode and bound the probability of the bad event $\overline{\cG}_\ell$ occurring.

\begin{theorem}[Regret without margin condition]
\label{thm:regret-without-margin}
Under Assumptions~\ref{assmpt:bounded}--\ref{assmpt:sparse-pd}, the regret of Algorithm~\ref{algo:private-lasso} is given by:
\begin{eqnarray*}
\cR(T) &\le&  \frac{1048(1+\log 4s_1)^{3/2}}{C_p^{2}}(\sigma^2+\sigma^2_{\cB}) (K-1) C^4_{1}C^2_r C^2_\cX s_1 \sqrt{T}\\
&+& 2(K-1)C_\theta C_\cX [64d\left(\sigma^2+\sigma^2_{\cB}\right)C^2_\cX s^2_1\frac{1}{\lambda^2_0}\log T + \frac{2}{C^2_0} \log T]\\
&=& \cO(\tfrac{s^{3/2}_0 \sqrt{T}\log^2 T \log\frac{\log T}{\delta}}{\varepsilon})
\end{eqnarray*}
\end{theorem}

%% file: paper/lower-bound.tex
In this section, we focus on private minimax risk estimation. To define the minimax risk we consider: 
\begin{enumerate}
\item $\{f_\theta,\ \theta \in \Theta \}$ is a family of statistical models supported over $\cX$ 
\item $X = \{X_1, X_2, \ldots, X_n\}$ are iid samples drawn from $f_{\theta^\ast}$ for some unknown $\theta^\ast \in \Theta$ and $\cM: \cX^n \times \Theta$ is an estimator of $\theta$
\item The loss function $\ell: \Theta \times \Theta \rightarrow \mR_{+}$ 
\end{enumerate}

The minimax risk is given by $\mE\left[\ell(\cM)(\cX),\theta)\right]$ is the risk associated with the estimator $\cM(\cX)$. The minimax risk is defined as: 
\[
\cR = \inf_{\cM}\sup_{\theta \in \Theta} \mE\left[\ell(\cM(\cX),\theta)\right]
\]
We restrict the set of estimators to the set of possible mechanisms $\cM_{\varepsilon,\delta}$ which are $(\varepsilon,\delta)$-DP.  In this section, we establish lower bounds for the differentially-private LASSO bandit. Our main contribution is to establish lower bounds on private estimation of the Fisher information matrix. This combined with ideas from~\cite{bassily2015local} leads to a private lower bound. To this end, let $\cP$ denote the family of distributions supported on $\cX$ and $\theta: \Theta \rightarrow $ denote its estimate. Let $\cA(\cX): \cX^{T} \rightarrow \Theta$ denote a mechanism which belongs to $\cA_{\varepsilon,\delta}$, the set of all $(\varepsilon,\delta)$-differentially private mechanisms. We measure the performance of $\hat{\theta}$ using the metric norm $\Vert \cdot \Vert_2$ and let $\ell: \mR^{+} \rightarrow \mR^{+}$ be an increasing function. Define the minimax risk: 
\[
\cR^{\ast} = \inf_{\cA \in \cA_{\varepsilon,\delta}} \sup_{P \in \cP} \mE\left[ \ell(\cA, \theta) \right]
\]

$\cR^\ast$ characterizes the worst case performance over all possible differentially private mechanisms $\cA_{\varepsilon,\delta}$. Our lower bound arguments are based on establishing a lower bound on the Bayesian risk of the estimation problem. Standard arguments such as those in~\cite{goldenshluger2013linear} eschew dimension dependence and privacy considerations. Instead, we consider a tracing-attack-based argument as in~\cite{cai2020cost} to obtain the correct dependence on these parameters. 

\begin{theorem}[Lower Bound on regret under JDP]
\label{thm:lower-bound}
For $\varepsilon >0$ and $T \ge 1$, the regret of any algorithm $\cA$ under Assumptions~\ref{assmpt:bounded}-~\ref{assmpt:sparse-pd} is given by:	
\[ 
\cR(\cA_{\varepsilon,\delta}) \ge \sigma^{2}\log T \min\left\{ s_0 \log d, \frac{s_0\log d}
{\varepsilon} \right\}
\]
\end{theorem}